\documentclass[5p,twocolumn]{elsarticle}

\usepackage[hidelinks]{hyperref}

\usepackage{lmodern}
\usepackage[T1]{fontenc}

\usepackage{caption} 
\usepackage[none]{hyphenat} 
\usepackage{array}

\newcolumntype{^}{>{\currentrowstyle}}

\usepackage{booktabs}
\usepackage{multirow}
\usepackage{threeparttable}

\usepackage{graphicx} 
\usepackage{float} 
\usepackage{subfig}
\graphicspath{{./}}
\usepackage{tikz}
\usetikzlibrary{shapes.geometric,arrows,matrix,positioning}
\usepackage{pgfplots}

\pgfplotsset{compat=newest} 
\newcounter{plotno}

\usetikzlibrary{pgfplots.groupplots,backgrounds}
\usepgfplotslibrary{units}

\tikzstyle{io} = [rectangle, minimum width=2cm, minimum height=2cm, text width=2cm, text centered]
\tikzstyle{block} = [rectangle, minimum width=2cm, minimum height=2cm, text width=2cm, text centered, draw=black, line width=0.3mm]
\tikzstyle{arrow_fc} = [line width=0.8mm,->,>=stealth]

\tikzstyle{sum} = [circle, minimum width=0.4cm, draw=black]
\tikzstyle{node1} = [circle, minimum width=0.4cm, draw=black, fill=brown!180]
\tikzstyle{node2} = [circle, minimum width=0.3cm, draw=black, fill=brown!180]
\tikzstyle{node3} = [circle, minimum width=0.2cm, draw=black, fill=brown!180]
\tikzstyle{leaf1} = [rectangle, minimum width=0.2cm, minimum height=0.2cm, draw=black, fill=green]
\tikzstyle{leaf2} = [rectangle, minimum width=0.2cm, minimum height=0.2cm, draw=black, fill=red]
\tikzstyle{arrow} = [line width=0.4mm,-]
\tikzstyle{arrow2} = [line width=0.6mm,->, >=stealth, color=cyan]
\tikzstyle{arrow3} = [line width=0.5mm,->, >=stealth,shorten >= 0.2cm, shorten <= 0.1cm]

\usepackage{mhchem} 
\usepackage{xfrac} 

\usepackage{setspace}

\usepackage{tikz}
\usepackage[upright]{fourier}
\usepackage{tkz-kiviat,numprint} 

\usetikzlibrary{decorations.pathreplacing, arrows, fit}
\usepackage{graphicx}
\usepackage{url}
\usepackage{booktabs}
\usepackage{textcomp}
\usepackage{epstopdf}
\usepackage{hhline}
\usepackage{xmpmulti}
\usepackage{siunitx}
\usepackage{adjustbox}

\pgfdeclarelayer{background}
\pgfdeclarelayer{foreground}
\pgfsetlayers{background,main,foreground}

\newcommand{\LegendBox}[3][]{%
\xdef\fitbox{}%
\coordinate[#1] (LegendBox_anchor) at (#2) ;
    \foreach \col/\item [count=\hi from 0] in {#3} {
       \node[color = \col,draw,
             fill  = \col!10,
             minimum width  = 5 ex,
             minimum height = 2 ex,
             name=b\hi,
       ] at ([yshift=\hi*4 ex,xshift=3ex]LegendBox_anchor) {};
       \node[anchor=west,xshift=1ex] at (b\hi.east) (c\hi) {\item};
       \xdef\fitbox{\fitbox(c\hi)}
   }%
 \node [draw,fit=\fitbox(LegendBox_anchor)] {};
}

\journal{Expert Systems with Applications}

\biboptions{authoryear}

\usepackage{amsthm,xparse}
\ExplSyntaxOn
\NewDocumentCommand{\delim}{ O{lr} m m }
 {
  \mickg_delim:nnn { #1 } { #2 } { #3 }
 }

\tl_new:N \l_mickg_lsize_tl
\tl_new:N \l_mickg_rsize_tl
\tl_new:N \l_mickg_ldel_tl
\tl_new:N \l_mickg_rdel_tl

\cs_new_protected:Npn \mickg_delim:nnn #1 #2 #3
 {
  \str_case:nnF { #1 }
   {
    { b }{\tl_set:Nn \l_mickg_lsize_tl {\big} \tl_set:Nn \l_mickg_rsize_tl {\big}}
    { B }{\tl_set:Nn \l_mickg_lsize_tl {\Big} \tl_set:Nn \l_mickg_rsize_tl {\Big}}
    { x }{\tl_set:Nn \l_mickg_lsize_tl {\bigg} \tl_set:Nn \l_mickg_rsize_tl {\bigg}}
    { X }{\tl_set:Nn \l_mickg_lsize_tl {\Bigg} \tl_set:Nn \l_mickg_rsize_tl {\Bigg}}
    { lr }{\tl_set:Nn \l_mickg_lsize_tl {\left} \tl_set:Nn \l_mickg_rsize_tl {\right}}
    { bB }{\tl_set:Nn \l_mickg_lsize_tl {\big} \tl_set:Nn \l_mickg_rsize_tl {\Big}}
    { Bb }{\tl_set:Nn \l_mickg_lsize_tl {\Big} \tl_set:Nn \l_mickg_rsize_tl {\big}}
    { xb }{\tl_set:Nn \l_mickg_lsize_tl {\bigg} \tl_set:Nn \l_mickg_rsize_tl {\big}}
    { xB }{\tl_set:Nn \l_mickg_lsize_tl {\bigg} \tl_set:Nn \l_mickg_rsize_tl {\Big}}
    { bx }{\tl_set:Nn \l_mickg_lsize_tl {\big} \tl_set:Nn \l_mickg_rsize_tl {\bigg}}
    { Bx }{\tl_set:Nn \l_mickg_lsize_tl {\Big} \tl_set:Nn \l_mickg_rsize_tl {\bigg}}
    { bX }{\tl_set:Nn \l_mickg_lsize_tl {\big} \tl_set:Nn \l_mickg_rsize_tl {\Bigg}}
    { BX }{\tl_set:Nn \l_mickg_lsize_tl {\Big} \tl_set:Nn \l_mickg_rsize_tl {\Bigg}}
    { xX }{\tl_set:Nn \l_mickg_lsize_tl {\bigg} \tl_set:Nn \l_mickg_rsize_tl {\Bigg}}
    { Xb }{\tl_set:Nn \l_mickg_lsize_tl {\Bigg} \tl_set:Nn \l_mickg_rsize_tl {\big}}
    { XB }{\tl_set:Nn \l_mickg_lsize_tl {\Bigg} \tl_set:Nn \l_mickg_rsize_tl {\Big}}
    { Xx }{\tl_set:Nn \l_mickg_lsize_tl {\Bigg} \tl_set:Nn \l_mickg_rsize_tl {\bigg}}
   }
   { \tl_clear:N \l_mickg_lsize_tl \tl_clear:N \l_mickg_rsize_tl }
  \str_case:nnF { #2 }
   {
    { s }{\tl_set:Nn \l_mickg_ldel_tl {[} \tl_set:Nn \l_mickg_rdel_tl {]}}
    { r }{\tl_set:Nn \l_mickg_ldel_tl {(} \tl_set:Nn \l_mickg_rdel_tl {)}}
    { b }{\tl_set:Nn \l_mickg_ldel_tl {\lbrace} \tl_set:Nn \l_mickg_rdel_tl {\rbrace}}
    { v }{\tl_set:Nn \l_mickg_ldel_tl {|} \tl_set:Nn \l_mickg_rdel_tl {|}}
    { a }{\tl_set:Nn \l_mickg_ldel_tl {\langle} \tl_set:Nn \l_mickg_rdel_tl {\rangle}}
    { dv }{\tl_set:Nn \l_mickg_ldel_tl {\|} \tl_set:Nn \l_mickg_rdel_tl {\|}}
    { rs }{\tl_set:Nn \l_mickg_ldel_tl {(} \tl_set:Nn \l_mickg_rdel_tl {]}}
    { sr }{\tl_set:Nn \l_mickg_ldel_tl {[} \tl_set:Nn \l_mickg_rdel_tl {)}}
    { rb }{\tl_set:Nn \l_mickg_ldel_tl {(} \tl_set:Nn \l_mickg_rdel_tl {\rbrace}}
    { sb }{\tl_set:Nn \l_mickg_ldel_tl {[} \tl_set:Nn \l_mickg_rdel_tl {\rbrace}}
    { br }{\tl_set:Nn \l_mickg_ldel_tl {\lbrace} \tl_set:Nn \l_mickg_rdel_tl {)}}
    { bs }{\tl_set:Nn \l_mickg_ldel_tl {\lbrace} \tl_set:Nn \l_mickg_rdel_tl {]}}
    { ra }{\tl_set:Nn \l_mickg_ldel_tl {(} \tl_set:Nn \l_mickg_rdel_tl {\rangle}}
    { sa }{\tl_set:Nn \l_mickg_ldel_tl {[} \tl_set:Nn \l_mickg_rdel_tl {\rangle}}
    { ba }{\tl_set:Nn \l_mickg_ldel_tl {\lbrace} \tl_set:Nn \l_mickg_rdel_tl {\rangle}}
    { ar }{\tl_set:Nn \l_mickg_ldel_tl {\langle} \tl_set:Nn \l_mickg_rdel_tl {)}}
    { as }{\tl_set:Nn \l_mickg_ldel_tl {\langle} \tl_set:Nn \l_mickg_rdel_tl {]}}
    { ab }{\tl_set:Nn \l_mickg_ldel_tl {\langle} \tl_set:Nn \l_mickg_rdel_tl {\rbrace}}
    { rv }{\tl_set:Nn \l_mickg_ldel_tl {(} \tl_set:Nn \l_mickg_rdel_tl {|}}
    { sv }{\tl_set:Nn \l_mickg_ldel_tl {[} \tl_set:Nn \l_mickg_rdel_tl {|}}
    { bv }{\tl_set:Nn \l_mickg_ldel_tl {\lbrace} \tl_set:Nn \l_mickg_rdel_tl {|}}
    { av }{\tl_set:Nn \l_mickg_ldel_tl {\langle} \tl_set:Nn \l_mickg_rdel_tl {|}}
    { vr }{\tl_set:Nn \l_mickg_ldel_tl {|} \tl_set:Nn \l_mickg_rdel_tl {)}}
    { vs }{\tl_set:Nn \l_mickg_ldel_tl {|} \tl_set:Nn \l_mickg_rdel_tl {]}}
    { vb }{\tl_set:Nn \l_mickg_ldel_tl {|} \tl_set:Nn \l_mickg_rdel_tl {\rbrace}}
    { va }{\tl_set:Nn \l_mickg_ldel_tl {|} \tl_set:Nn \l_mickg_rdel_tl {\rangle}}
    { rdv }{\tl_set:Nn \l_mickg_ldel_tl {(} \tl_set:Nn \l_mickg_rdel_tl {\|}}
    { sdv }{\tl_set:Nn \l_mickg_ldel_tl {[} \tl_set:Nn \l_mickg_rdel_tl {\|}}
    { bdv }{\tl_set:Nn \l_mickg_ldel_tl {\lbrace} \tl_set:Nn \l_mickg_rdel_tl {\|}}
    { adv }{\tl_set:Nn \l_mickg_ldel_tl {\langle} \tl_set:Nn \l_mickg_rdel_tl {\|}}
    { vdv }{\tl_set:Nn \l_mickg_ldel_tl {|} \tl_set:Nn \l_mickg_rdel_tl {\|}}
    { dvr }{\tl_set:Nn \l_mickg_ldel_tl {\|} \tl_set:Nn \l_mickg_rdel_tl {)}}
    { dvs }{\tl_set:Nn \l_mickg_ldel_tl {\|} \tl_set:Nn \l_mickg_rdel_tl {]}}
    { dvb }{\tl_set:Nn \l_mickg_ldel_tl {\|} \tl_set:Nn \l_mickg_rdel_tl {\rbrace}}
    { dva }{\tl_set:Nn \l_mickg_ldel_tl {\|} \tl_set:Nn \l_mickg_rdel_tl {\rangle}}
    { dvv }{\tl_set:Nn \l_mickg_ldel_tl {\|} \tl_set:Nn \l_mickg_rdel_tl {|}}
    { lr }{\tl_set:Nn \l_mickg_ldel_tl {(} \tl_clear:N \l_mickg_rdel_tl}
    { rr }{\tl_set:Nn \l_mickg_ldel_tl {)} \tl_clear:N \l_mickg_rdel_tl}
    { ls }{\tl_set:Nn \l_mickg_ldel_tl {[} \tl_clear:N \l_mickg_rdel_tl}
    { rs }{\tl_set:Nn \l_mickg_ldel_tl {]} \tl_clear:N \l_mickg_rdel_tl}
    { lb }{\tl_set:Nn \l_mickg_ldel_tl {\lbrace} \tl_clear:N \l_mickg_rdel_tl}
    { rb }{\tl_set:Nn \l_mickg_ldel_tl {\rbrace} \tl_clear:N \l_mickg_rdel_tl}
    { la }{\tl_set:Nn \l_mickg_ldel_tl {\langle} \tl_clear:N \l_mickg_rdel_tl}
    { ra }{\tl_set:Nn \l_mickg_ldel_tl {\rangle} \tl_clear:N \l_mickg_rdel_tl}
    { ov }{\tl_set:Nn \l_mickg_ldel_tl {|} \tl_clear:N \l_mickg_rdel_tl}
    { odv }{\tl_set:Nn \l_mickg_ldel_tl {\|} \tl_clear:N \l_mickg_rdel_tl}
    { ssi }{\tl_set:Nn \l_mickg_ldel_tl {[} \tl_set:Nn \l_mickg_rdel_tl {\mathclose{[}}}
    { sis }{\tl_set:Nn \l_mickg_ldel_tl {\mathopen{]}} \tl_set:Nn \l_mickg_rdel_tl {]}}
    { sisi }{\tl_set:Nn \l_mickg_ldel_tl {\mathopen{]}} \tl_set:Nn \l_mickg_rdel_tl {\mathclose{[}}}
   }
   { \tl_set:Nn \l_mickg_ldel_tl {(} \tl_set:Nn \l_mickg_rdel_tl {)} } 
  \tl_if_empty:NTF \l_mickg_rdel_tl
   {
    \l_mickg_lsize_tl \l_mickg_ldel_tl #3
   }
   {
    \l_mickg_lsize_tl \l_mickg_ldel_tl #3 \l_mickg_rsize_tl \l_mickg_rdel_tl
   }
 }
\ExplSyntaxOff

\begin{document}

\begin{frontmatter}

\title{Supervised Learning in Automatic Channel Selection for Epileptic Seizure Detection}

\author{Nhan Truong}
\author{Levin Kuhlmann}
\author{Mohammad Reza Bonyadi}
\author{Jiawei Yang}
\author{Andrew Faulks}
\author{Omid Kavehei}

\begin{abstract}
Detecting seizure using brain neuroactivations recorded by intracranial electroencephalogram (iEEG) has been widely used for monitoring, diagnosing, and closed-loop therapy of epileptic patients, however, computational efficiency gains are needed if state-of-the-art methods are to be implemented in implanted devices. We present a novel method for automatic seizure detection based on iEEG data that outperforms current state-of-the-art seizure detection methods in terms of computational efficiency while maintaining the accuracy. The proposed algorithm incorporates an automatic channel selection (ACS) engine as a pre-processing stage to the seizure detection procedure. The ACS engine consists of supervised classifiers which aim to find iEEG channels which contribute the most to a seizure. Seizure detection stage involves feature extraction and classification. Feature extraction is performed in both frequency and time domains where spectral power and correlation between channel pairs are calculated. Random Forest is used in classification of interictal, ictal and early ictal periods of iEEG signals. Seizure detection in this paper is retrospective and patient-specific. iEEG data is accessed via Kaggle, provided by International Epilepsy Electro-physiology Portal. The dataset includes a training set of $6.5$ hours of interictal data and $41$~minin ictal data and a test set of $9.14$ hours. Compared to the state-of-the-art on the same dataset, we achieve $49.4\%$ increase in computational efficiency and $400$~mins better in average for detection delay. The proposed model is able to detect a seizure onset at $91.95\%$ sensitivity and $94.05\%$ specificity with a mean detection delay of $2.77$ s. The area under the curve ($AUC$) is $96.44\%$, that is comparable to the current state-of-the-art with $AUC$ of $96.29\%$.  
\end{abstract}


\end{frontmatter}


\section{Introduction}

Epileptic seizure affects nearly $1\%$ of global population but only two thirds can be treated by medicine and approximately $7-8\%$ can be cured by surgery \citep{Litt2002}. Therefore, seizure onset detection and subsequent seizure suppression becomes important for the patients that cannot be cured by neither drug nor surgery. Early detection can allow early electrical stimulation to suppress the seizure \citep{Echauz2007}. In this paper, we focus on how to effectively and reliably detect seizure onset based on iEEG patterns. Causes and treatment of seizure is beyond the scope of this paper.

EEG has been commonly used in brain-computer interface thanks to the convenient real-time readings and high temporal resolution of EEG signals \citep{Zeng2015,ZhangH2013}. In recent years, EEG has provided a promising possibility to detect and even predict an epileptic seizure \citep{tieng2016mouse,Fatichah2014, Parvez2015,Saab2005,Osorio2009,kuhlmann2009}. For seizure detection, \citet{Fatichah2014} used a combination of principle component analysis (PCA) and neural network with fuzzy membership function that can achieve accuracy rate up to $97.64\%$ . \citet{tieng2016mouse} combined wavelet de-noising with adapted Continuous Wavelet Transform in their algorithm and were able to achieve sensitivity of $96.72\%$ and specificity of $94.69\%$ with EEG data from mice. Another remarkable method is to transform EEG signals into images so as to leverage image processing techniques \citep{Parvez2015}. This approach was able to obtain $98.91\%$ sensitivity and $94.35\%$ specificity. \citet{Zabihi2016} reconstructed EEG phase spaces using time-delay embedding method and PoinCare section. The phase spaces were then reduced by PCA before being fed to linear discriminant analysis (LDA) and Naive Bayesian classifiers. This approach achieved $88.27\%$ sensitivity and $93.21\%$ specificity in seizure detection.

\citet{shoeb2009application} deployed $8$ filters spanning the frequency range of $0.5$--$24$~Hz for each $2$-s EEG epoch of all channels, then concatenated $3$ epochs to form a feature set to be fed to a SVM classifier. This approach was tested with the CHB-MIT EEG dataset and was able to detect $96\%$ of $163$ test seizures with a mean detection delay of $4.6$ seconds. Using the same CHB-MIT dataset, EEG signal was transformed into an image representation using $2$-D projection of the patient electrodes and the magnitude of $3$ different frequency bands spanning the range of $0$--$49$~Hz of each $1$~s block of EEG signal \citep{thodoroff2016learning}. The recurrent convolutional neural network took $30$ consecutive blocks as inputs to perform feature extraction and classification. The patient-specific detectors in this method have comparable performance compared to the proposed method by \citet{shoeb2009application}.

\begin{table*}[htbp]
\normalsize

\caption{Summary of existing EEG-based seizure detection methods \label{tbl:summary}}
\resizebox{1.0\textwidth}{!}{
\begin{tabular}{ l c c c c c c c l c c }
\toprule
\multirow{2}{*}{Reference} & \multirow{2}{*}{EEG type} & \multirow{2}{3.1em}{\centering No. of patients} & \multirow{2}{3.1em}{No. of seizures} & \multicolumn{2}{c}{Data duration} & \multirow{2}{3.2em}{\centering Patient\\-specific} & \multirow{2}{5.5em}{\centering Split data for training} & \multirow{2}{3.5em}{\centering Testing\\sensitivity} & \multirow{2}{3.1em}{\centering FDR$^\ast$} & \multirow{2}{5.5em}{\begin{tabular}{@{}c@{}} Mean\\detection delay \end{tabular}} \\
& & & & ictal & interictal \\ 
\toprule  \hline

\citet{Saab2005}& scalp & $44$ & $195$ & \multicolumn{2}{c}{$1012$ h$^\dagger$} & No & $64\%$ & $76\%$ & $0.34$/h & $9.8$s \\ 
\citet{kuhlmann2009}& scalp & $21$ & $88$ & \multicolumn{2}{c}{$525$ h$^\dagger$} & No & $70\%$ & $81\%$ & $0.60$/h & $16.9$s \\ 
\citet{Wang2016}& scalp & $10$ & $44$ & $72$~min & $121$ h & Yes & $80\%$ & $91.44\%$ & $99.34\%$ & n/a \\ 
\citet{Zabihi2016}& scalp & $24$ & $161$ & $2.55$ h & $169$ h & Yes & $25\%$ & $88.27\%$ & $93.21\%$ & n/a \\

\citet{Fatichah2014} & intracranial$^\ddagger$ & n/a & n/a & $39.3$~min & $2.62$ h & n/a & $90\%$ & $94.55\%$ & $98.41\%$ & n/a \\
\citet{Hills2014}& intracranial & $12$ & $48$ & $41$~min & $6.5$ h & Yes & $50\%$ & $91.33\%$ & $94.02\%$ & $3.17$s \\ 
\citet{Parvez2015}& intracranial & $21$ & $87$ & $58$ h & $490$ h & n/a & $80\%$ & $100\%$ & $97\%$ & n/a \\ 
\bottomrule 

\end{tabular}}{\scriptsize
\begin{tablenotes}
\item[] {$^\ast$~False detection rate (FDR) or specificity.}
\item[] {$^\dagger$~Duration of ictal and interictal were not provided separately.}
\item[] {$^\ddagger$~Intracranial EEG for seizure class and both intracranial and extracranial for non-seizure class.}
\end{tablenotes}}
\end{table*}

Prominent feature extraction techniques consider characteristics in both frequency and time domain. As an efficient tool for time-frequency-energy analysis, wavelet-based filters were used to extract a ratio of seizure content of the short foreground in comparison with the background \citep{Saab2005,Osorio2009}. \citet{Saab2005} applied Bayes' formula on extracted features to estimate the probability of seizure in EEG signals. This method achieved an impressively short onset detection delay of $9.8$~s with $76\%$ sensitivity and $0.34$/h false positive rate. \citet{kuhlmann2009} extended Saab and Gotman's method by combining extra features to find a superior detector. Their method was able to achieve a sensitivity of $81\%$, a false positive rate of $0.60$/h, and a median detection delay of $16.9$~s on a dataset of $525$ h of scalp EEG data.

\begin{figure*}[htbp]
\centering

\resizebox{0.8\textwidth}{!}{
\includegraphics{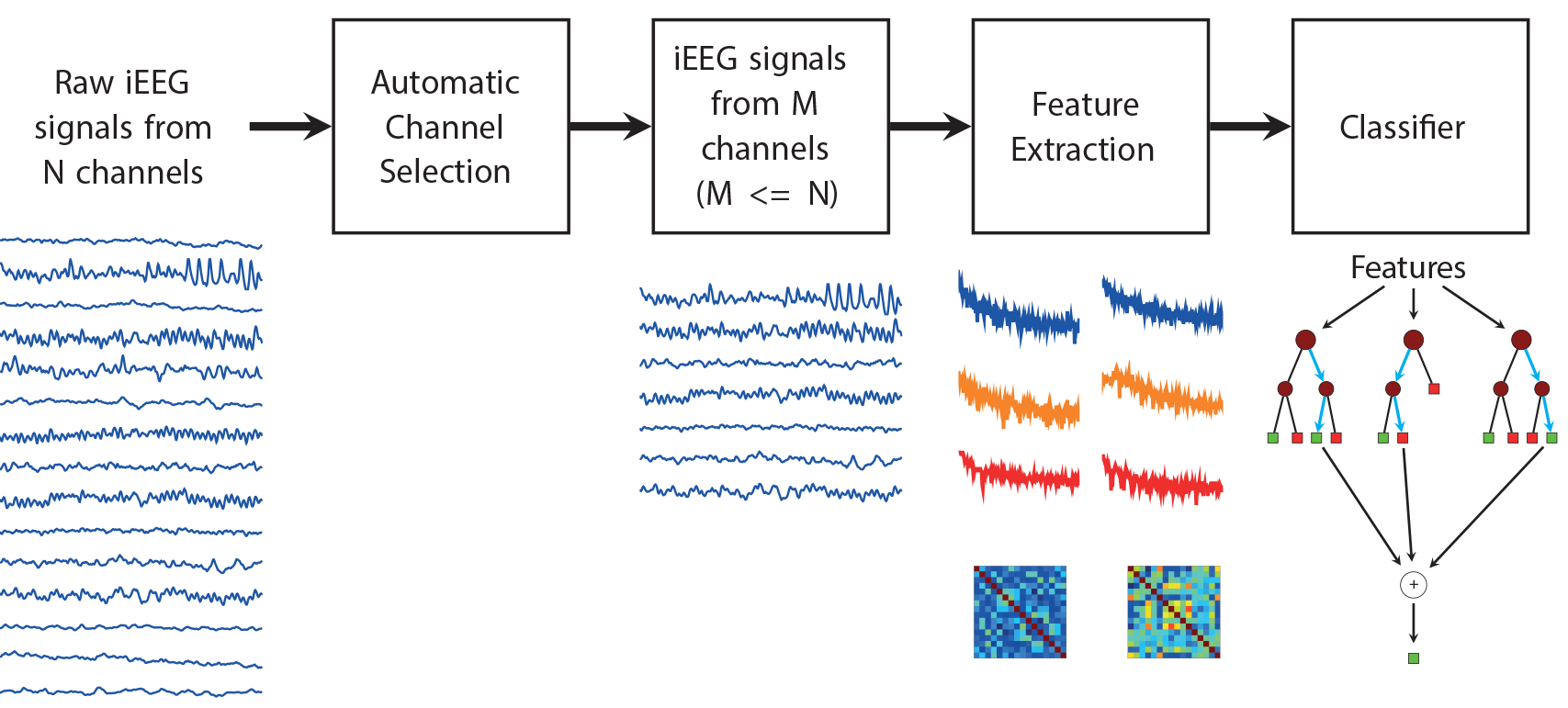}
}
\caption{Flowchart of the proposed method. Raw iEEG data from all $N$ channels is fed to ACS to find $M$ channels which contribute the most to a seizure. The ACS engine is executed one time only for each subject at the beginning and indexes of the $M$ channels are stored on hard-disk. Feature extraction in both frequency and time domains is done on the $M$ channels. Extracted features are fed to a classifier using Random Forest algorithm to discriminate interictal, ictal and early ictal epochs.}
\label{fig:flow}
\end{figure*}

The current state-of-the-art seizure detection method proposed by \citet{Hills2014} for the dataset considered here is implemented and extended in this paper. The dataset is derived from a Kaggle seizure detection competition in which \citet{Hills2014} scored $AUC$ of $96.29\%$ and announced as the winner. Description of the dataset is provided in Section~\ref{sub:sec:dataset}. In this paper, we significantly enhanced computational efficiency of Hill's method by employing an automatic channel selection algorithm. This enabled us to process data as accurately with reduced number of channels. Table~\ref{tbl:summary} summarizes the existing EEG-based seizure detection methods in recent years.

The remainder of this paper is organized as follows. In Section~\ref{sec:propmeth}, after describing the dataset, we propose automatic channel selection engine that helps to reduce the number of channels to be processed. This section also presents spatio-temporal feature extraction and Random Forest classifier used for seizure detection. Section~\ref{sec:eval} evaluates the performance of the proposed model with comparison against the state-of-the-art method on the same dataset. Section~\ref{sec:discuss} concludes the achievement of the paper.

\section{Proposed method}\label{sec:propmeth}

The intracranial EEG data was recorded on multiple subjects with varying number of channels and sampling rates. We propose an automatic channel selection engine to filter out channels which are less relevant to seizure. The engine accepts the raw iEEG data, their corresponding labels, and the number of channels to be selected, $M$, and determines indexes of channels that are most relevant for seizure detection. Indexes of these $M$ channels are stored on hard-disk so the engine only needs to be executed one time at the beginning for each subject. Feature extraction was performed in both frequency and time domain on the selected channels. Information extracted in frequency and time domains was concatenated and fed to a Random Forest classifier. Fig.~\ref{fig:flow} presents flowchart of the proposed method.

\subsection{Dataset} \label{sub:sec:dataset}

Dataset being analyzed in this paper is obtained from \citet{Kaggle}. Intracranial EEG signals were recorded from $4$ dogs and $8$ patients with epileptic seizures. Recordings were sampled at $400$~Hz from $16$ electrodes for dogs, and sampled at $500$~Hz or $5$~kHz from varying number of electrodes (ranging from $16$ to $72$) for humans. The data was pre-organized into $1$~s iEEG epochs annotated as ictal for seizure states or interictal for seizure-free states. Interictal data was captured not less than one hour before or after a seizure onset and randomly chosen from the recorded data. Each ictal segment also came with the time in seconds between the seizure onset and first data point of the segment. The training dataset is consisted of $41$~min of ictal data and $6.5$~hours of interictal data. Summary of the training dataset is presented in Table~\ref{tbl:dataset}. Note that early ictal state in this paper is the ictal state occurring within the first $15$~s from the seizure onset. The proposed method was tested with a hidden dataset provided by Kaggle. This dataset consists of $9.14$~hours of unlabeled iEEG data \citep{Kaggle}.

\begin{table}[htbp]
\normalsize
\centering
\caption{Summary of the dataset\label{tbl:dataset}}
\resizebox{0.47\textwidth}{!}{
\begin{tabular}{  lccccc  }
\toprule
	\multirow{3}{*}{Subject} & \multirow{3}{4em}{\centering No. of electrodes} & \multirow{3}{4.5em} {\centering Ictal data length~(s)} & \multirow{3}{4.5em}{\centering Interictal data length~(s)} & \multirow{3}{5.5em}{\centering Unlabeled data length~(s)} & \multirow{3}{4em}{\centering Train/Test ratio} \\ \\ \\
\toprule  \hline
	Dog--1 & $16$ & $178$ & $418$ & $3181$ & $0.19$ \\ 
	Dog--2 & $16$ & $172$ & $1148$ & $2997$ & $0.44$ \\ 
	Dog--3 & $16$ & $480$ & $4760$ & $4450$ & $1.18$ \\ 
	Dog--4 & $16$ & $257$ & $2790$ & $3013$ & $1.01$ \\ 
	Patient--1 & $68$ & $70$ & $104$ & $2050$ & $0.08$ \\ 
	Patient--2 & $16$ & $151$ & $2990$ & $3894$ & $0.81$ \\ 
	Patient--3 & $55$ & $327$ & $714$ & $1281$ & $0.81$ \\ 
	Patient--4 & $72$ & $20$ & $190$ & $543$ & $0.39$ \\ 
	Patient--5 & $64$ & $135$ & $2610$ & $2986$ & $0.92$ \\ 
	Patient--6 & $30$ & $225$ & $2772$ & $2997$ & $1$ \\ 
	Patient--7 & $36$ & $282$ & $3239$ & $3601$ & $0.98$ \\ 
	Patient--8 & $16$ & $180$ & $1710$ & $1922$ & $0.98$ \\ 
\midrule
	\textbf{Total} & & \boldmath$2477$ & \boldmath$23445$ & \boldmath$32915$ & \boldmath$0.79$\\ 
\bottomrule
\end{tabular}}
\end{table}

\subsection{Automatic channels selection}
The intracranial EEG data was recorded using various number of channels ($16$, $30$, $36$, $55$, $64$, $68$, $72$). Large number of channels yields higher computational complexity as it requires more data to be analyzed. This can also deteriorate the diversity of iEEG data, hence degrade the performance of seizure detection, because some channels may capture irrelevant information \citep{Guyon2003}. One can leverage bio-medical knowledge to manually select which channels genuinely contribute to the seizure. However, it is hard, if not impossible, to disclose a set of channels that are significant for all subjects. It is required to use the expertise to analyze every subject (or group of subjects) to proclaim a list of significant channels with regards to each subject (or group of subjects) which is manifestly a time-consuming task.

We propose a novel approach for automatic channels selection (ACS) as follows. The labeled data is first transformed to obtain frequency information. Specifically, FFT is applied onto the raw iEEG data on all $N$ channels. FFT values are then sliced to extract data in $1$-Hz bins in the range of $1$--$30$~Hz. $\log_{10}$ is then applied to the magnitudes. The transformed data is a $N \times 30$~minatrix where $30$ is the number of $1$-Hz bins in the range of $1$--$30$~Hz. If the channels correlation is involved in ACS stage, it will be confusing to identify which channels are the most important based on the importance level of the correlation between each pair of channels. Therefore, the correlation among channels is disregarded in this stage. Each individual channel becomes a feature to be fed to classifiers. One or a set of classifiers determine the importance level of each feature or channel. There are several options of classifiers using different ensemble algorithms such as Gradient Boosting, AdaBoost and Random Forest. If multiple classifiers are used, the final importance level of each channel is the sum of importance values obtained from all classifiers. The measure of feature importance in this paper is implemented using scikit-learn ensemble library \citep{scikit}. The importance of a feature is estimated by how often that feature is used in split points of each individual decision tree of the ensemble classifier \citep{scikit}. It is important to note that only train dataset was involved in the ACS stage.

The output of the channel selection algorithm is a set of $M$ channels sorted based on the level of their contribution to the detection of a seizure. In this paper, we selected the value of $M$ through some experiments aiming at maximising the final $AUC$ score. This selection, however, could be automated by setting a threshold on the importance value of the channels.

\subsection{Feature extraction}

\subsubsection{Feature extraction in frequency domain}
\begin{figure}[htbp]
\centering
\resizebox{0.47\textwidth}{!}{
	\includegraphics{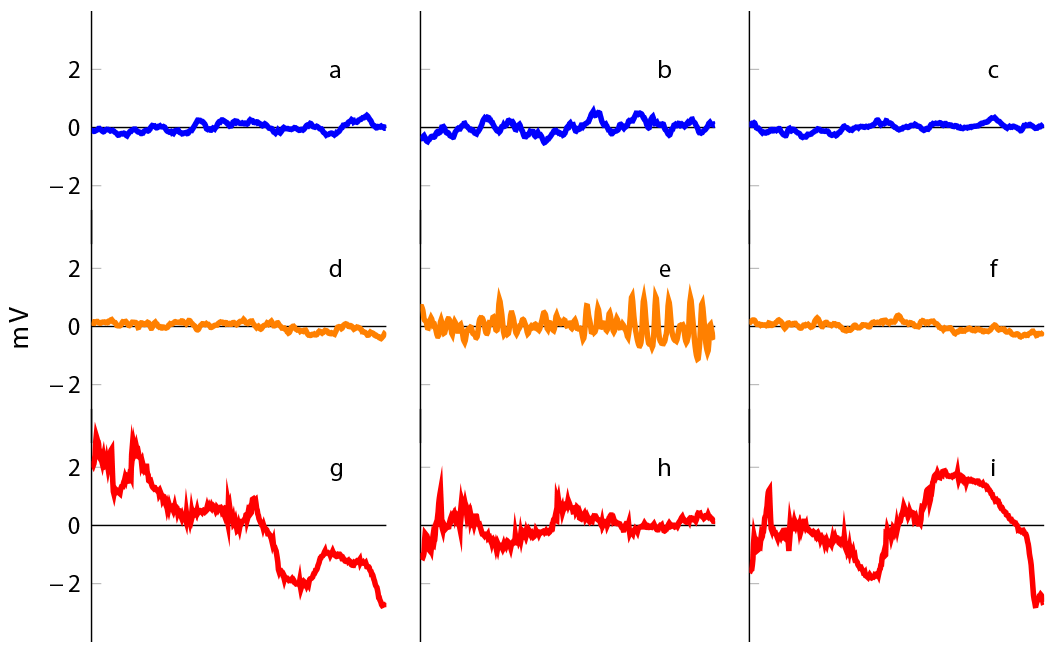}
}

\caption{Sample $1$~s iEEG recordings. (a, b and c) interictal; (d, e and f) ictal at early state (within $15$~s from seizure onset); (g, h and i): ictal after early state. iEEG signals presented in one column, (e.g. a, d and g) are recorded from the same channel.}
\label{fig:sample1}
\end{figure}

\begin{figure}[htbp]
\vspace{-1.27\baselineskip}
\centering
\resizebox{0.48\textwidth}{!}{
	\includegraphics{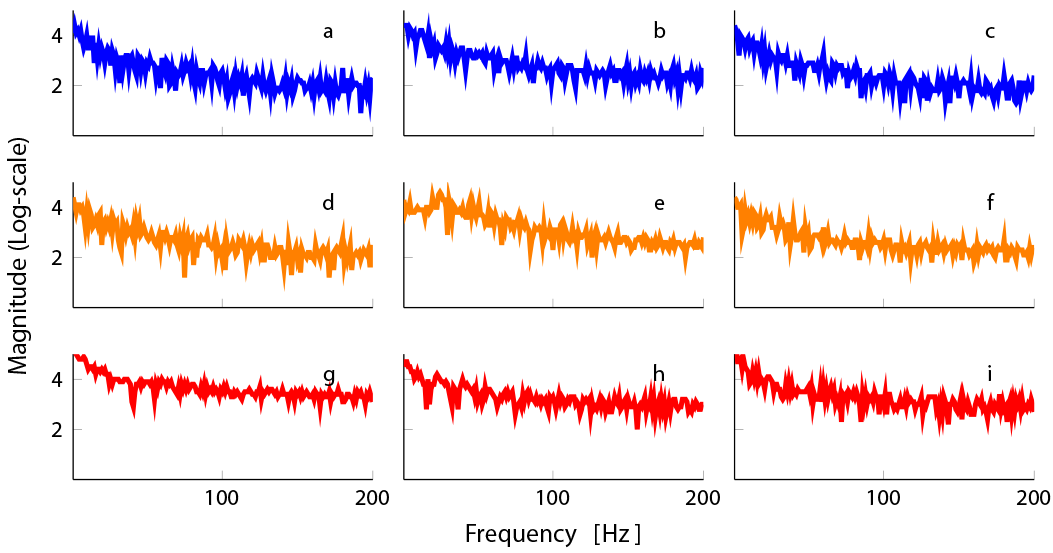}
}

\caption{Sample $1$~s iEEG recordings power spectrum. (a, b and c) interictal; (d, e and f) ictal at early state; (g, h and i): ictal after early state. iEEG signals presented in one column, (e.g. a, d and g) are recorded from the same channel. Subplots in this figure are one-by-one associated with subplots in Fig.~\ref{fig:sample1}.}	  \label{fig:sample2}
\end{figure}

The iEEG signals from $M$ selected channels are transformed by FFT. The transformed data then is filtered to discard high frequency noise and low frequency artifacts. Frequency range of $1$--$47$~Hz was shown to achieve the best performance for the dataset \citep{Hills2014}. Eigenvalues have been used as an effective technique to discriminate ictal epochs in \citep{Zhang2016,Hills2014,Sardouie2015}. In order to compute eigenvalues, spectral power is primarily normalized (zero mean and standard deviation of one) along each channel before estimating cross spectral matrix \citep{Hills2014}. Contrary to the Hills' feature extraction, we did not use cross spectral coefficients as a feature because our empirical observation shows that such feature could worsen detection accuracy. Sample recordings and corresponding power spectrum for ictal and interictal segments of Patient--1 are illustrated in Figs.~\ref{fig:sample1} and \ref{fig:sample2}.

The feature set in frequency domain consists of:
\begin{itemize}
	\item [--] {Spectral power in $1$~Hz bins in range of $1$--$47$~Hz by applying $\log_{10}$ to the magnitude of FFT transformation, and}
	\item [--] {Eigenvalues, sorted in descending order, of cross spectral matrix on all selected channels of the above spectral power.}
\end{itemize}

\subsubsection{Feature extraction in time domain}
\begin{figure}[htbp]
\captionsetup[subfloat]{captionskip=-1pt}
\vspace{-1.27\baselineskip}
\centering
\subfloat[\label{subfig:interictal}]
{\includegraphics[height=0.12\textwidth]{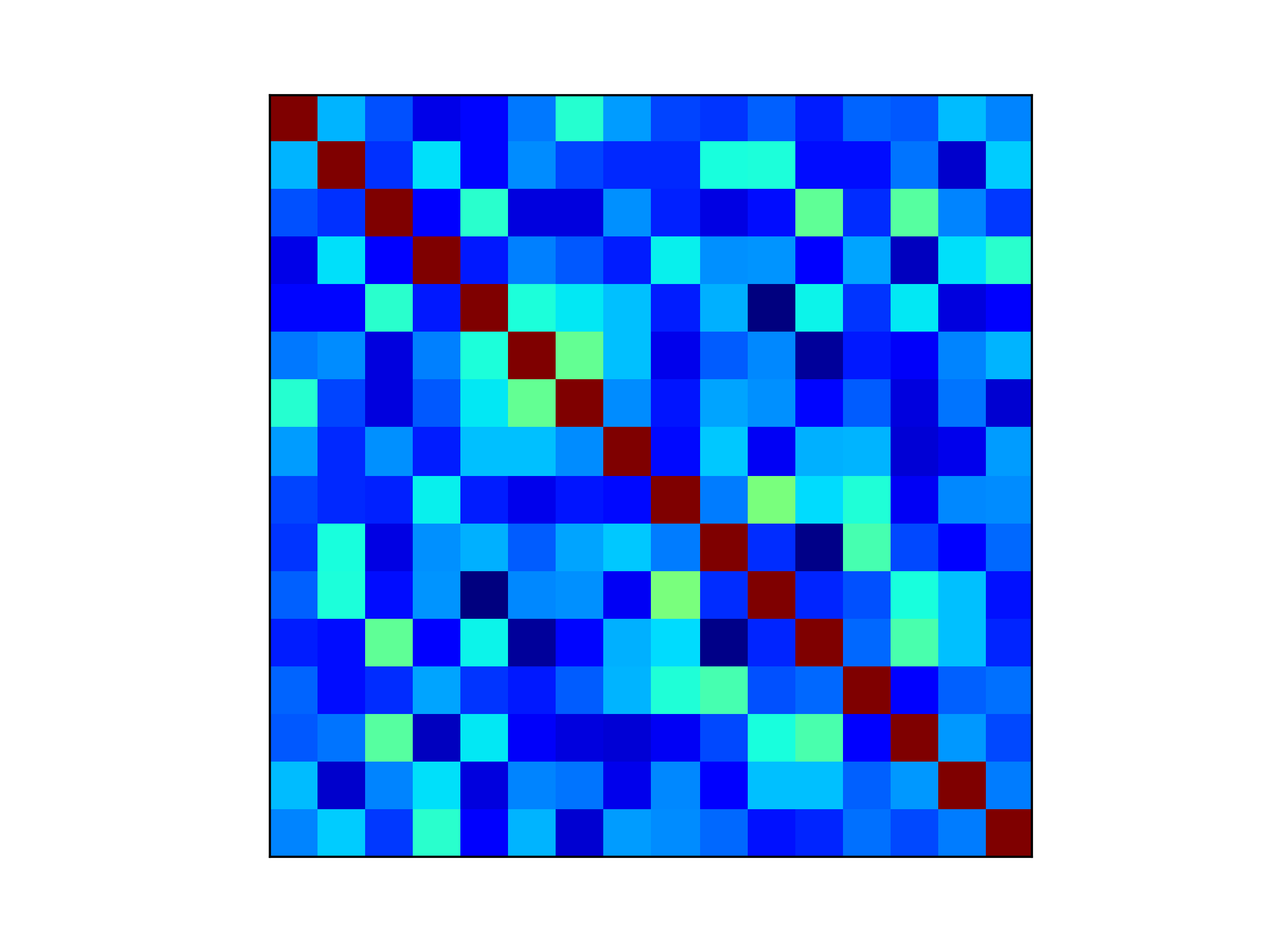}}
\hfill
\subfloat[\label{subfig:early}]
{\includegraphics[height=0.12\textwidth]{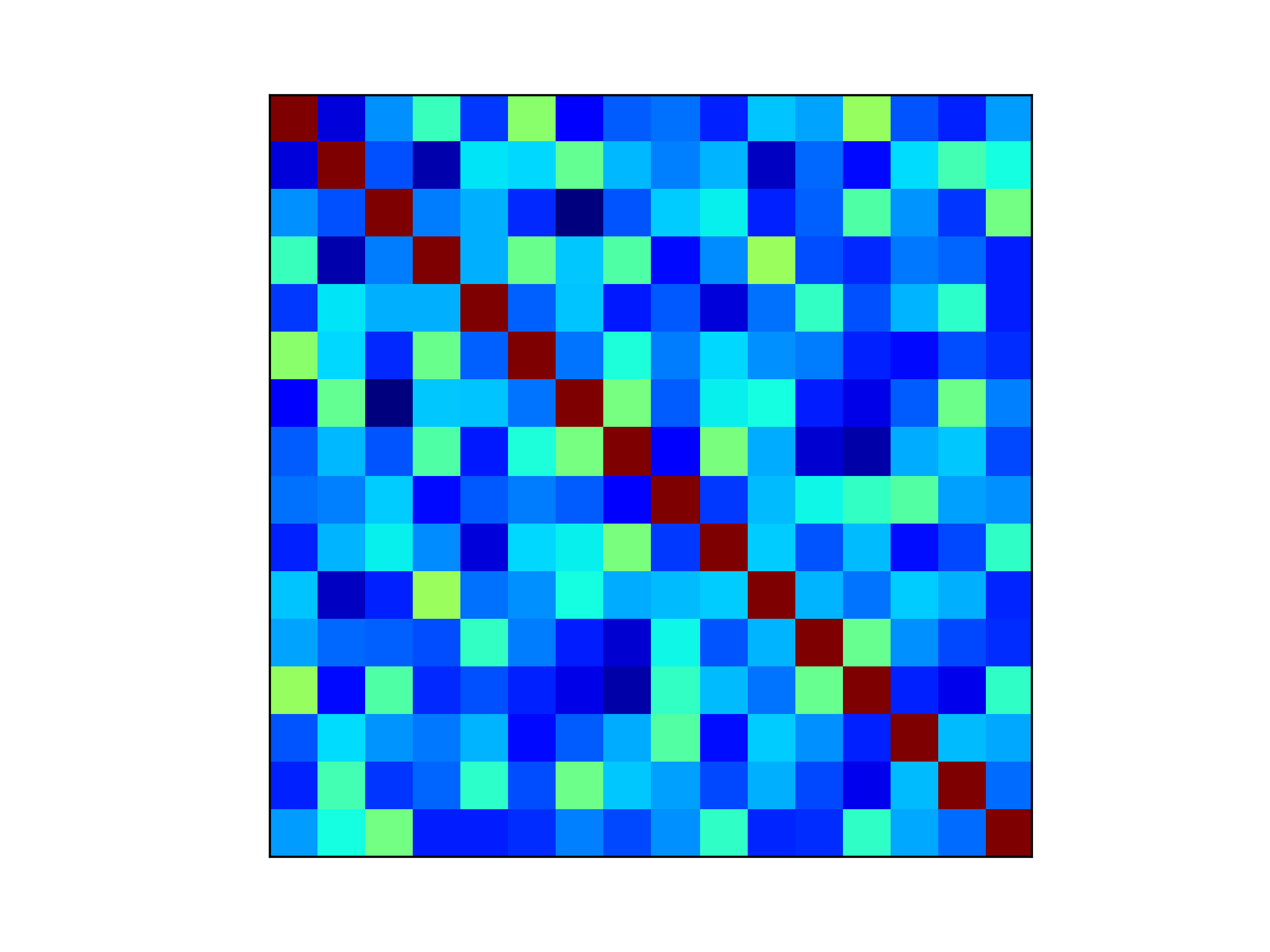}}	
\hfill
\subfloat[\label{subfig:ictal}]
{\includegraphics[height=0.12\textwidth]{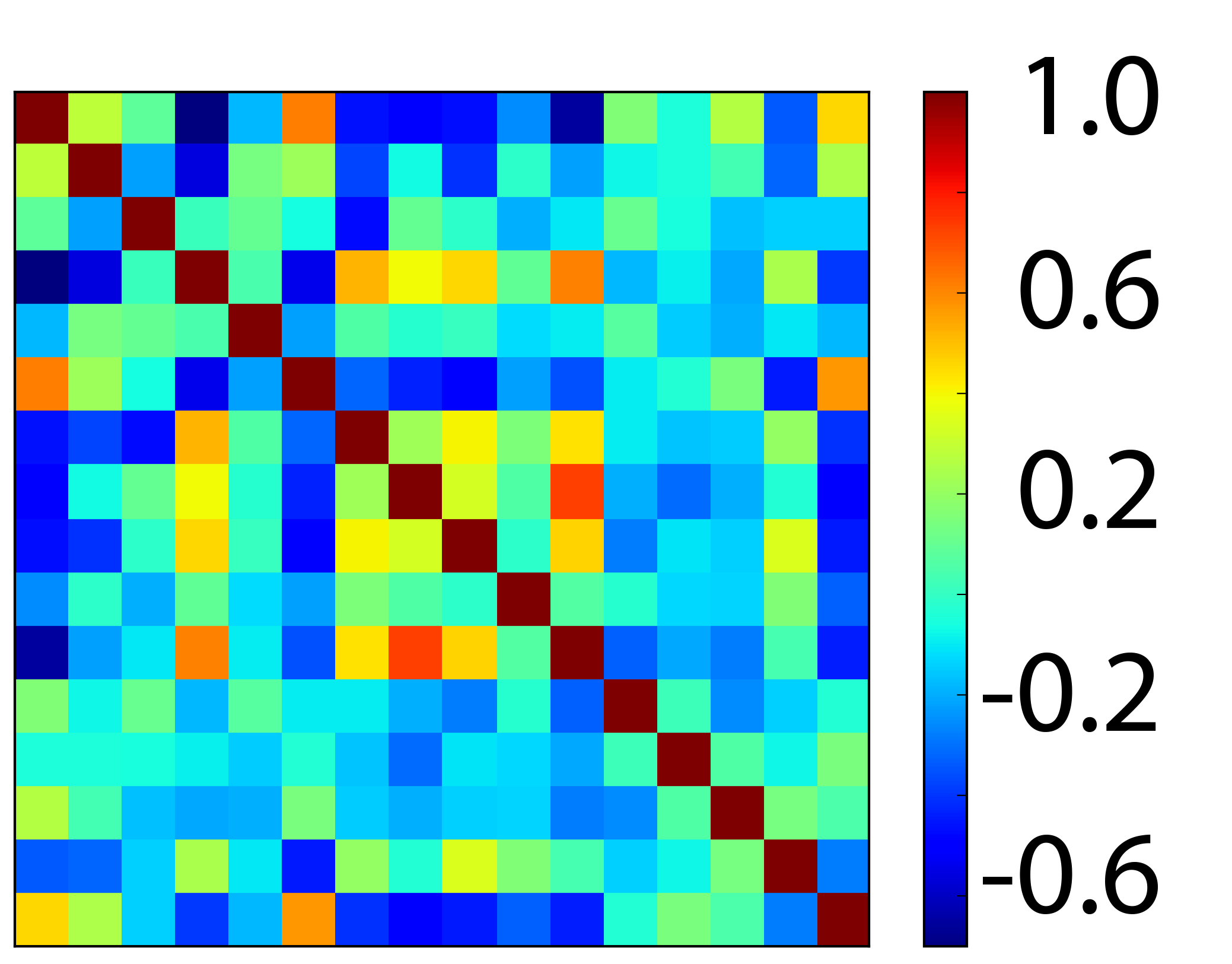}}	
\caption{Covariance matrix: a) interictal; b) ictal at early state; c) ictal. Correlation between channels is very low in interictal period. The channels are more correlated after the seizure onset and highly correlated in ictal state. \label{fig:cov}}
\end{figure}

Raw iEEG signals are firstly re-sampled to $400$~Hz. Similarly to frequency domain, filtered iEEG data is normalized to zero mean and unity standard deviation along each channel prior to computing covariance matrix and its eigenvalues. As illustrated in Fig.~\ref{fig:cov}, iEEG data from $16$ selected channels of Patient--1 have a very low correlation to each others in interictal states. The correlation slightly increases when seizure is at early state and becomes remarkable beyond the early state.

The feature set in time domain consists of:
\begin{itemize}
	\item [--] {Coefficients in upper triangle of correlation matrix of iEEG signals from selected channels, and}
	\item [--] {Eigenvalues of the correlation matrix above, sorted in descending order.}
\end{itemize}

\subsection{Classifier}
Random Forest algorithm was first proposed by \citet{Breiman2001}. The algorithm uses a large set of decision trees to acquire an average results. Random Forest has been shown with good performance on dataset with high dimensional datasets in biology and medical fields \citep{Scornet2016,Huynh2016,Cabezas2016}. This paper will not go in deep about its mathematical properties as they can be found in \citep{Breiman2001,Scornet2016} but rather on fine-tuning the parameters to achieve the highest performance with the given feature sets.

Random Forest classifier in this paper is implemented using scikit-learn library \citep{scikit}. Parameters of the classifier are reused from the approach proposed by \citet{Hills2014} with $3000$ decision trees. The classifier analyses each $1$~s iEEG epoch and categorizes them into $3$ classes as outputs: early ictal (ictal within $15$~s from the onset), ictal, and interictal. Regarding sensitivity and specificity evaluation, the Random Forest classifier is adjusted from three-class classifier to binary classifier which detects whether a $1$~s iEEG signal is ictal or interictal.

\begin{table*}[htp]
\centering
\caption{Comparison between state-of-the-art and proposed method on computational efficiency. ACS engine with $M$=$16$ for all subjects. \label{tbl:computational}}
\resizebox{0.93\textwidth}{!}{
\begin{tabular}{ l c c c c c c c c}

\toprule
	\ & \ & \ & \multicolumn{2}{c}{\citet{Hills2014}}  & \multicolumn{3}{c}{Proposed method} & \   \\ 
\cmidrule(lr){4-5}
\cmidrule(lr){6-8}
	\multirow{2}{4em}{\centering Subject} & \multirow{2}{4em}{\centering No. of electrodes} & \multirow{2}{7em}{\centering Data duration \\(min)} & \multirow{2}{3.0em}{\centering FA$^\dagger$ (s)} & \multirow{2}{5em}{\centering Training (s)} & \multirow{2}{3.5em}{\centering ACS$^\ast$ (s)} & \multirow{2}{3.0em}{\centering FA$^\dagger$ (s)} & \multirow{2}{5em}{\centering Training (s)} & \multirow{2}{8em}{\centering Processing time improvement} \\ 
	& & & & & & & &   \\ 
\toprule  \hline
	
	Dog--1 & $16$ & $9.9$ & $2$ & $23.7$ & n/a & $1.7$ & $23.7$ & n/a  \\ 
	Dog--2 & $16$ & $22$ & $3.8$ & $62.7$ & n/a & $3.5$ & $62.5$ & n/a  \\ 
	Dog--3 & $16$ & $87.3$ & $15.4$ & $341.8$ & n/a & $13.8$ & $348.8$ & n/a  \\ 
	Dog--4 & $16$ & $50.8$ & $8.8$ & $144.9$ & n/a & $8.5$ & $149.2$ & n/a \\ 
	Patient--1 & $68$ & $2.9$ & $2.7$ & $14.1$ & $6.5$ & $0.7$ & $13.1$ & $17.9\%$  \\ 
	Patient--2 & $16$ & $52.4$ & $21.3$ & $105.8$ & n/a & $22.3$ & $106.3$ & n/a \\ 
	Patient--3 & $55$ & $17.4$ & $25.5$ & $80.7$ & $48.4$ & $6$ & $57.6$ & $40.1\%$  \\ 
	Patient--4 & $72$ & $3.5$ & $7.1$ & $13.3$ & $10.6$ & $1.5$ & $11.6$ & $35.8\%$  \\ 
	Patient--5 & $64$ & $45.8$ & $79.1$ & $519.3$ & $180.3$ & $29.4$ & $162.4$ & $67.9\%$  \\ 
	Patient--6 & $30$ & $50$ & $38.5$ & $146.2$ & $93.5$ & $15.2$ & $98.1$ & $38.7\%$  \\ 
	Patient--7 & $36$ & $58.7$ & $53.7$ & $435.3$ & $171.9$ & $19$ & $244$ & $46.2\%$  \\ 
	Patient--8 & $16$ & $31.5$ & $13.3$ & $59.7$ & n/a & $13.4$ & $60.1$ & n/a  \\

\midrule
	\textbf{Average} & \ & \ &  &  &  &  &  & \boldmath$41.1\%$ \\
\bottomrule

\end{tabular}}{\scriptsize
\begin{tablenotes}
\scriptsize
\item[] {~~~~~~~~~~$^\ast$~Automatic channel selection (ACS) time.}
\item[] {~~~~~~~~~~$^\dagger$~Feature extraction (FA) time.}
\end{tablenotes}}	
\end{table*}

\section{Evaluation}\label{sec:eval}

In this section, we will compare the efficacy of the proposed method with the current state-of-the-art method proposed by \cite{Hills2014} on the same dataset. Metrics used to test the proposed approach are area under the receiver operating characteristic curve ($AUC$), sensitivity, specificity and onset detection delay with $2$--fold cross-validation. In the first fold, a half of the seizures per subject were chosen as training set, the rest were for validation. Note that each seizure only appears in either the training set or the validation set. The interictal data was divided randomly into two equal length sets, one for training and one for validation. In the second fold, roles of training and validation sets in the first fold were swapped. Final results reported in the following are the average of outputs from the two folds. We will also test with the hidden dataset consisting of $9.14$ hours of unlabeled iEEG data to compare the $AUC$ scores computed by the leader board of Kaggle for the two methods.

\begin{table}[htbp]
\centering
\caption{$AUC$ Comparison between state-of-the-art and proposed method with $M$=$16$ for all subjects.\label{tbl:comparison}}
\resizebox{0.46\textwidth}{!}{
\begin{tabular}{ l c c c c c c }

	\toprule
	& \multicolumn{3}{c}{\citet{Hills2014}}  & \multicolumn{3}{c}{Proposed method} \  \\ 
\cmidrule(lr){2-4}
\cmidrule(lr){5-7}
	\multirow{2}{3em}{Subject} & \multirow{2}{3em}{\centering $AUC_{E}$ (\%)} & \multirow{2}{3em}{\centering$AUC_{S} $ (\%)} & \multirow{2}{3em}{\centering$AUC$ (\%)} & \multirow{2}{3em}{\centering$AUC_{E}$ (\%)} & \multirow{2}{3em}{\centering$AUC_{S}$ (\%)} & \multirow{2}{3em}{\centering$AUC$ (\%)} \ \\ 
& & & \\	
	\toprule \hline
	
	Dog--1 & $97.59$ & $99.40$ & $98.49$ & $97.14$ & $99.11$ & $98.13$ \\ 
	Dog--2 & $95.00$ & $98.85$ & $96.92$ & $94.02$ & $97.76$ & $95.89$ \\ 
	Dog--3 & $96.77$ & $99.52$ & $98.14$ & $96.57$ & $99.47$ & $98.02$ \\ 
	Dog--4 & $99.87$ & $97.16$ & $98.51$ & $99.88$ & $97.08$ & $98.48$ \\ 
	Patient--1 & $90.94$ & $98.14$ & $94.54$ & $96.05$ & $99.00$ & $97.52$ \\ 
	Patient--2 & $99.29$ & $99.34$ & $99.31$ & $99.15$ & $99.32$ & $99.23$ \\ 
	Patient--3 & $87.98$ & $94.25$ & $91.11$ & $87.63$ & $92.69$ & $90.16$ \\ 
	Patient--4 & $100$ & $100$ & $100$ & $99.63$ & $99.63$ & $99.63$ \\ 
	Patient--5 & $83.02$ & $89.38$ & $86.20$ & $87.54$ & $90.73$ & $89.13$ \\ 
	Patient--6 & $98.61$ & $99.80$ & $99.20$ & $98.86$ & $99.83$ & $99.35$ \\ 
	Patient--7 & $89.97$ & $96.02$ & $92.99$ & $93.84$ & $97.58$ & $95.71$ \\ 
	Patient--8 & $81.46$ & $97.82$ & $89.64$ & $79.83$ & $97.80$ & $88.81$
 \\ 

\midrule	
	\textbf{Average} & \boldmath$93.37$ & \boldmath$97.47$ & \boldmath$95.42$ & \boldmath$94.18$ & \boldmath$97.50$ & \boldmath$95.84$ \\

	\bottomrule

\end{tabular}}
\end{table}

\begin{table}[htbp]
\centering
\caption{Comparison between state-of-the-art and proposed method with $M$=$16$ for all subjects on sensitivity (SEN), specificity (SEP) and onset detection delay with corresponding threshold (Thres.) for binary classification of seizure and non-seizure states.\label{tbl:sen_spe_delay}}
\resizebox{0.49\textwidth}{!}{
\begin{tabular}{ l c c c c c c c c  }
\toprule
	\ &  \multicolumn{4}{c}{\citet{Hills2014}}  & \multicolumn{4}{c}{Proposed method} \    \\ 
\cmidrule(lr){2-5}
\cmidrule(lr){6-9}
	\multirow{2}{2.5em}{\centering Subject} & \multirow{2}{2em}{\centering Delay (s)} & \multirow{2}{2em}{\centering SEN (\%)} & \multirow{2}{2em}{\centering SPE (\%)} & \multirow{2}{2em}{\centering Thres.} & \multirow{2}{2em}{\centering Delay (s)} & \multirow{2}{2em}{\centering SEN (\%)} & \multirow{2}{2em}{\centering SPE (\%)} & \multirow{2}{2em}{\centering Thres.} \\ 
& & & & \\
\toprule \hline
	
	Dog--1 & $1.92$ & $95.83$ & $97.13$ & $0.34$ & $1.92$ & $94.46$ & $98.09$ & $0.42$ \\
Dog--2 & $2.25$ & $92.86$ & $89.29$ & $0.11$ & $2.25$ & $92.44$ & $91.82$ & $0.13$ \\
Dog--3 & $1.83$ & $95.84$ & $97.33$ & $0.14$ & $2.17$ & $94.80$ & $98.28$ & $0.21$ \\
Dog--4 & $1$ & $89.55$ & $89.90$ & $0.09$ & $1$ & $89.55$ & $90.26$ & $0.09$ \\
Patient--1 & $4$ & $92.31$ & $96.16$ & $0.36$ & $3$ & $94.87$ & $96.16$ & $0.28$ \\
Patient--2 & $2$ & $95.03$ & $99.20$ & $0.29$ & $2$ & $95.03$ & $99.20$ & $0.28$ \\
Patient--3 & $1.75$ & $85.42$ & $76.61$ & $0.17$ & $1.38$ & $86.67$ & $77.45$ & $0.13$ \\
Patient--4 & $1$ & $100$ & $98.95$ & $0.25$ & $1$ & $100$ & $97.37$ & $0.23$ \\
Patient--5 & $10$ & $71.67$ & $87.78$ & $0.05$ & $3.75$ & $77.78$ & $83.49$ & $0.03$ \\
Patient--6 & $1.75$ & $98.24$ & $98.85$ & $0.19$ & $2.25$ & $97.35$ & $99.21$ & $0.26$ \\
Patient--7 & $6$ & $86.51$ & $99.14$ & $0.14$ & $2$ & $99.59$ & $99.79$ & $0.20$ \\
Patient--8 & $4.5$ & $92.78$ & $97.90$ & $0.22$ & $4.5$ & $92.78$ & $97.90$ & $0.20$ \\

\midrule
	\textbf{Average} & \boldmath$3.17$ & \boldmath$91.33$ & \boldmath$94.02$ & \ & \boldmath$2.27$ & \boldmath$92.94$ & \boldmath$94.08$ & \ \\
\bottomrule

\end{tabular}}
\end{table}

This paper aims to detect whether a given $1$~s iEEG segment represents a seizure and whether that segment is within the first $15$~s (early) of its respective seizure. The metric for performance evaluation is the average of the two $AUC$s of the two detections~\cite{Kaggle}, and is given by

\begin{equation}
AUC={\frac{1}{2}}(AUC_{\rm S} + AUC_{\rm E}),
\end{equation}
where,
\begin{itemize}
	\item [--] {$AUC_{\rm S}$ is $AUC$ for two classes: ictal (including early seizure) and interictal, and}
	\item [--] {$AUC_{\rm E}$ is $AUC$ for two classes: early seizure and non-early-seizure (including ictal states after $15$~s from onset and interictal states).}
\end{itemize}

Table~\ref{tbl:computational} summarizes the gain in processing time, including time for feature extraction and classifier training for each subject. Since we chose $M$=$16$ in ACS engine, subjects with number of electrodes less than or equal to $16$ would skip the channel selection stage; hence, the processing time of the proposed method is comparable with that of the state-of-the-art method for these subjects.  Overall gain in computational efficiency for the patients with more than $16$ iEEG channels is $41.1\%$. Therefore, the automatic channel selection is promising for real-time seizure detection application.

Table~\ref{tbl:comparison} summarizes cross-validation $AUC$ scores using state-of-the-art and proposed methods for each subject. Overall cross-validation $AUC$ scores of the two methods are comparable. The proposed method was tested with the hidden dataset and acquired an $AUC$ score at $96.58\%$ which is comparable with the state-of-the-art's score at $96.29\%$ \citep{Hills2014}.

\begin{table*}[htp]
\centering
\caption{Comparison between state-of-the-art and proposed method on computational efficiency. ACS engine with $M$ was optimized on the training data per subject.\label{tbl:computational2}}
\resizebox{0.97\textwidth}{!}{
\begin{tabular}{ l c c c c c c c c c }

\toprule
	\ & \ &\ & \ & \multicolumn{2}{c}{\citet{Hills2014}}  & \multicolumn{3}{c}{Proposed method} & \   \\ 
\cmidrule(lr){5-6}
\cmidrule(lr){7-9}
	\multirow{2}{4em}{\centering Subject} & \multirow{2}{4em}{\centering No. of electrodes} & \multirow{2}{2em}{\centering $M$} & \multirow{2}{7em}{\centering Data duration \\(min)} & \multirow{2}{3em}{\centering FA$^\dagger$ (s)} & \multirow{2}{4em}{\centering Training (s)} & \multirow{2}{3em}{\centering ACS$^\ast$ (s)} & \multirow{2}{3em}{\centering FA$^\dagger$ (s)} & \multirow{2}{4em}{\centering Training (s)} & \multirow{2}{8em}{\centering Processing time improvement} \\ 
	& & & & & & & & &   \\ 
\toprule \hline
	
	Dog--1 & $16$ & $9$ & $9.9$ & $2$ & $23.7$ & $6$ & $1.5$ & $19.8$ & $17.1\%$  \\ 
	Dog--2 & $16$ & $10$ & $22$ & $3.8$ & $62.7$ & $15$ & $2.8$ & $49.9$ & $20.8\%$  \\ 
	Dog--3 & $16$ & $8$ & $87.3$ & $15.4$ & $341.8$ & $77$ & $9.7$ & $224.5$ & $34.4\%$  \\ 
	Dog--4 & $16$ & $13$ & $50.8$ & $8.8$ & $144.9$ & $36$ & $7.1$ & $122.2$ & $15.9\%$ \\ 
	Patient--1 & $68$ & $16$ & $2.9$ & $2.7$ & $14.1$ & $5$ & $0.7$ & $11.7$ & $26.2\%$  \\ 
	Patient--2 & $16$ & $11$ & $52.4$ & $21.3$ & $105.8$ & $39$ & $11.2$ & $82.8$ & $26.0\%$ \\ 
	Patient--3 & $55$ & $8$ & $17.4$ & $25.5$ & $80.7$ & $31$ & $4$ & $45.1$ & $53.8\%$  \\ 
	Patient--4 & $72$ & $4$ & $3.5$ & $7.1$ & $13.3$ & $8$ & $0.8$ & $11.1$ & $41.7\%$  \\ 
	Patient--5 & $64$ & $16$ & $45.8$ & $79.1$ & $519.3$ & $115$ & $28.2$ & $158.1$ & $68.9\%$  \\ 
	Patient--6 & $30$ & $8$ & $50$ & $38.5$ & $146.2$ & $56$ & $9.6$ & $60.6$ & $62.0\%$  \\ 
	Patient--7 & $36$ & $13$ & $58.7$ & $53.7$ & $435.3$ & $118$ & $15.6$ & $198.6$ & $56.2\%$  \\ 
	Patient--8 & $16$ & $8$ & $31.5$ & $13.3$ & $59.7$ & $23$ & $5.4$ & $41.6$ & $35.6\%$  \\

\midrule
	\textbf{Average} & \ & \ & \ & \ & \ & \ & \ & \ & \boldmath$49.4\%$ \\
\bottomrule

\end{tabular}}
\begin{tablenotes}
\scriptsize
\item[] {$^\ast$~Automatic channel selection (ACS) time.}
\item[] {$^\dagger$~Feature extraction (FA) time.}
\end{tablenotes}	
\end{table*}

Regarding sensitivity and specificity evaluation, the Random Forest classifier is adjusted from three-class classifier to binary classifier which detects whether a $1$~s iEEG signal is ictal or interictal. Table~\ref{tbl:sen_spe_delay} describes the comparison between the state-of-the-art and proposed method on sensitivity, specificity and onset detection delay. The threshold of the classifier's output used to separate whether a $1$~s iEEG segment is ictal or interictal was determined per subject. The value of threshold was selected to achieve the balance between sensitivity and specificity (ie., the higher threshold value yields the higher specificity but the lower sensitivity and vice versa).

Proposed method achieved a comparable performance to the state-of-the-art in terms of sensitivity and specificity. However, the proposed method yields a considerable improvement in mean onset detection delay. Onset detection delay indicates the time in seconds after that the classifier can detect a seizure onset. Delay is $1$~s if the first $1$~s ictal iEEG segment at seizure onset can be correctly detected. Since iEEG signals are divided into $1$~s epochs, the minimum onset detection delay could be achieved is $1$ s. Table~\ref{tbl:sen_spe_delay} shows that the proposed model has shorter onset detection delay by $900$~mins than the current best method.

\section{Discussions}\label{sec:discuss}

We presented a seizure detection method based on a novel approach for automatic iEEG channel selection that provides comparable performance to the state-of-the-art method for the dataset considered. Although this leads to an extra overhead computing time in the beginning, the impact overall processing time is negligible because the channel selection need to be executed one time only for each subject. The advantages of the automatic channel selection, on the other hand, are remarkable. Firstly, redundant and unrelated iEEG signals are eliminated which helps to improve efficacy of seizure detection system. Secondly, since the amount of data to processed is reduced, the processing time is also reduced. Gain in computational complexity becomes visible and significant for subjects with large number of channels. For instance, by reducing number of channels to be analyzed from $72$ to $16$ for Patient--4 (see Table~\ref{tbl:computational}, processing time can be improved by $67.9\%$.

\begin{figure}[htbp]
\resizebox{0.45\textwidth}{!}{
\begin{tikzpicture}
\begin{scope}
 \tkzKiviatDiagram[
        radial=3,
        label style/.append style={font=\Large,text width=,align=center,shift={(24pt,0pt)}},
        radial  style/.style ={-},
        lattice style/.style ={blue!30},
        ]
      {Detection~delay, Processing~time, Number~of~channels}
 \tkzKiviatLine[thick,color=red,mark=ball,
                ball color=red,mark size=2.5pt,fill=red!20,opacity=.7](10,10,10)
 \tkzKiviatLine[thick,color=blue,mark=ball,
                mark size=2.5pt,fill=blue!20,opacity=.6](7,7,4.6) 
 \tkzKiviatLine[thick,color=green,mark=ball, ball color=green,
                mark size=2.5pt,fill=green!20,opacity=.5](8,5,2.8)
                \end{scope}

\LegendBox[shift={(-8cm,0.5cm)}]{current bounding box.south east}%
          {red/ {\Large State-of-the-art method},
           blue/ {\Large Proposed method with $M$=$16$},
           green/ {\Large Proposed method with optimized $M$}}     

\node[xshift=-4.5cm, yshift=-3cm] at (current bounding box.north east) {\Large Smaller triangle is better};

\end{tikzpicture}
}
\caption{Comparison between state-of-the-art and proposed method with two set of $M$ in terms of detection delay, number of processed channels and processing time. \label{tbl:seiz_detect_radar}}                      
\end{figure}
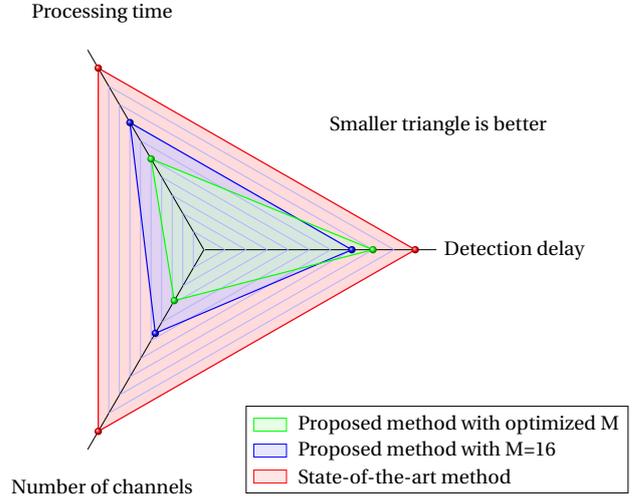

\begin{table}[htbp]
\centering
\caption{$AUC$ Comparison between state-of-the-art and proposed method with $M$ was optimized on the training data per subject.\label{tbl:comparisonAUC2}}
\resizebox{0.46\textwidth}{!}{
\begin{tabular}{lcccccc}

	\toprule
	& \multicolumn{3}{c}{\citet{Hills2014}}  & \multicolumn{3}{c}{Proposed method} \  \\ 
\cmidrule(lr){2-4}
\cmidrule(lr){5-7}
	\multirow{2}{3em}{Subject} & \multirow{2}{3em}{\centering $AUC_{E}$ (\%)} & \multirow{2}{3em}{\centering$AUC_{S} $ (\%)} & \multirow{2}{3em}{\centering$AUC$ (\%)} & \multirow{2}{3em}{\centering$AUC_{E}$ (\%)} & \multirow{2}{3em}{\centering$AUC_{S}$ (\%)} & \multirow{2}{3em}{\centering$AUC$ (\%)} \ \\ 
& & & \\	
	\toprule \hline
	
	Dog--1 & $97.59$ & $99.40$ & $98.49$ & $96.93$ & $99.11$ & $98.02$ \\ 
	Dog--2 & $95.00$ & $98.85$ & $96.92$ & $93.90$ & $97.56$ & $95.73$ \\ 
	Dog--3 & $96.77$ & $99.52$ & $98.14$ & $95.67$ & $99.17$ & $97.42$ \\ 
	Dog--4 & $99.87$ & $97.16$ & $98.51$ & $99.83$ & $97.45$ & $98.64$ \\ 
	Patient--1 & $90.94$ & $98.14$ & $94.54$ & $96.05$ & $99.00$ & $97.52$ \\ 
	Patient--2 & $99.29$ & $99.34$ & $99.31$ & $99.21$ & $99.30$ & $99.25$ \\ 
	Patient--3 & $87.98$ & $94.25$ & $91.11$ & $86.29$ & $93.31$ & $89.80$ \\ 
	Patient--4 & $100$ & $100$ & $100$ & $100$ & $100$ & $100$ \\ 
	Patient--5 & $83.02$ & $89.38$ & $86.20$ & $87.54$ & $90.73$ & $89.13$ \\ 
	Patient--6 & $98.61$ & $99.80$ & $99.20$ & $99.04$ & $99.89$ & $99.46$ \\ 
	Patient--7 & $89.97$ & $96.02$ & $92.99$ & $95.01$ & $97.92$ & $96.47$ \\ 
	Patient--8 & $81.46$ & $97.82$ & $89.64$ & $80.42$ & $98.23$ & $89.33$ \\ 

\midrule	
	\textbf{Average} & \boldmath$93.37$ & \boldmath$97.47$ & \boldmath$95.42$ & \boldmath$94.16$ & \boldmath$97.64$ & \boldmath$95.90$ \\

	\bottomrule

\end{tabular}}
\end{table}

Spectral power, correlation matrix and its eigenvalues on iEEG channels in both frequency and time domains have been shown as important features in seizure detection using iEEG recordings. The proposed subject-specific approach has a mean seizure onset detection delay of $2.27$~s that is critical, for example, for an electrical stimulator to suppress the seizure on time.

In order to further gain computational efficiency, the number of selected channels $M$ is optimized on the training data per subject based on cross-validation $AUC$. A range of number of channels, from $1$ to the total number of channels, is used to find the corresponding cross-validation $AUC$. $M$ is chosen to be the smallest with $AUC$ not less than $1\%$ compared to the best $AUC$. Using this approach, computational efficiency and mean detection delay are improved by $49.4\%$ and $400$~mins, respectively, compare to the state-of-the-art, see \citet{Hills2014}, while a comparable performance is maintained as demonstrated in Tables~\ref{tbl:computational2},~\ref{tbl:comparisonAUC2}, and \ref{tbl:sen_spe_delay2}. The overall $AUC$, when tested with the hidden test dataset, is $96.44\%$, comparable to that of the state-of-the-art at $96.29\%$. Fig.~\ref{tbl:seiz_detect_radar} demonstrates the advantages of the proposed method in terms of detection delay, number of channels to be analyzed and processing time.

\begin{table}[htbp]
\centering
\caption{Comparison between state-of-the-art and proposed method with $M$ was optimized on the training data per subject on sensitivity (SEN), specificity (SEP) and onset detection delay with corresponding threshold (Thres.) for binary classification of seizure and non-seizure states.\label{tbl:sen_spe_delay2}}
\resizebox{0.49\textwidth}{!}{
\begin{tabular}{ l c c c c c c c c  }
\toprule
	\ &  \multicolumn{4}{c}{\citet{Hills2014}}  & \multicolumn{4}{c}{Proposed method} \    \\ 
\cmidrule(lr){2-5}
\cmidrule(lr){6-9}
	\multirow{2}{2.5em}{\centering Subject} & \multirow{2}{2em}{\centering Delay (s)} & \multirow{2}{2em}{\centering SEN (\%)} & \multirow{2}{2em}{\centering SPE (\%)} & \multirow{2}{2em}{\centering Thres.} & \multirow{2}{2em}{\centering Delay (s)} & \multirow{2}{2em}{\centering SEN (\%)} & \multirow{2}{2em}{\centering SPE (\%)} & \multirow{2}{2em}{\centering Thres.} \\ 
& & & & \\
\toprule \hline
	
	Dog--1 & $1.92$ & $95.83$ & $97.13$ & $0.34$ & $2.5$ & $93.99$ & $96.41$ & $0.34$ \\
Dog--2 & $2.25$ & $92.86$ & $89.29$ & $0.11$ & $2.25$ & $92.44$ & $89.20$ & $0.11$ \\
Dog--3 & $1.83$ & $95.84$ & $97.33$ & $0.14$ & $2.5$ & $94.17$ & $97.25$ & $0.16$ \\
Dog--4 & $1$ & $89.55$ & $89.90$ & $0.09$ & $1$ & $92.16$ & $90.61$ & $0.09$ \\
Patient--1 & $4$ & $92.31$ & $96.16$ & $0.36$ & $3.5$ & $93.59$ & $97.12$ & $0.38$ \\
Patient--2 & $2$ & $95.03$ & $99.20$ & $0.29$ & $2$ & $95.03$ & $99.20$ & $0.26$ \\
Patient--3 & $1.75$ & $85.42$ & $76.61$ & $0.17$ & $1$ & $90.45$ & $76.19$ & $0.12$ \\
Patient--4 & $1$ & $100$ & $98.95$ & $0.25$ & $1$ & $100$ & $100$ & $0.49$ \\
Patient--5 & $10$ & $71.67$ & $87.78$ & $0.05$ & $4.75$ & $73.89$ & $85.94$ & $0.04$ \\
Patient--6 & $1.75$ & $98.24$ & $98.85$ & $0.19$ & $2.75$ & $96.48$ & $99.39$ & $0.33$ \\
Patient--7 & $6$ & $86.51$ & $99.14$ & $0.14$ & $5.5$ & $87.35$ & $99.14$ & $0.15$ \\
Patient--8 & $4.5$ & $92.78$ & $97.90$ & $0.22$ & $4.5$ & $93.89$ & $98.19$ & $0.26$ \\

\midrule
	\textbf{Average} & \boldmath$3.17$ & \boldmath$91.33$ & \boldmath$94.02$ & \ & \boldmath$2.77$ & \boldmath$91.95$ & \boldmath$94.05$ & \ \\
\bottomrule
	
\end{tabular}}
\end{table}

\section{Conclusion}
Detection of seizure, especially at its early state, is crucial for patients who cannot be treated by drugs or surgery. Precise seizure detection allows electrical stimulation to timely interrupt the alteration of consciousness and subsequent convulsions. Although high performing seizure detectors are available, translating state-of-the-art seizure detection methods into battery-saving hardware implementations in implantable seizure control devices requires greater gains in computational efficiency. This paper proposed automatic channels selection engine as a mechanism to adequately determine most informative iEEG recordings prior to feature extraction. The engine gave rise to significant computational efficiency improvements on subjects having large number of recording channels. For Patient--5, the computational efficiency was improved by $68.9\%$ (see Fig.~\ref{tbl:computational2}). The overall results of the proposed method were comparable with that of the state-of-the-art while it save $49.4\%$ of the processing time and reduced the mean detection delay by $400$~mins, both critical factors for real-world applications.

\section{Acknowledgement}
N. Truong greatly acknowledges The Commonwealth Scientific and Industrial Research Organisation (CSIRO) financial support via a PhD Scholarship, PN~50041400.

\section*{References}

\bibliography{mybibfile}

\end{document}